%% file: main.tex
\pdfoutput=1

\documentclass[11pt]{article}

\usepackage[]{naacl2021}

\usepackage{times}
\usepackage{latexsym}

\usepackage[T1]{fontenc}
\usepackage{environ}
\usepackage[utf8]{inputenc}
\usepackage{microtype}
\usepackage{booktabs}
\usepackage{multirow}
\usepackage{tabularx}
\usepackage{times}
\usepackage{url}
\usepackage{latexsym}
\usepackage{relsize}
\usepackage{wrapfig}
\usepackage{ragged2e}
\usepackage{graphicx}
\usepackage{arabtex}
\usepackage{arydshln}
\usepackage{url}
\usepackage{lipsum}  
\usepackage{enumitem}
\usepackage{amssymb}
\usepackage{amsmath}
\usepackage{utf8}
\usepackage{float}
\usepackage{diagbox}
\usepackage{footnote}
\makesavenoteenv{tabular}
\makesavenoteenv{table}
\usepackage{verbatim}
\usepackage[bottom]{footmisc}
\usepackage{tablefootnote}
\usepackage{tabularx}
\usepackage[toc,page]{appendix}
\usepackage{hyperref}
\usepackage{stackengine}
\usepackage{algorithm}
\newdimen\fontdim
\newdimen\upperfontdim
\newdimen\lowerfontdim
\newif\ifmoreiterations
\makeatletter
\NewEnviron{fitbox}[2]{
  \def\buildbox{%
    \setbox0\vbox{\hbox{\minipage{#1}%
      \fontsize{\fontdim}{1.2\fontdim}%
      \selectfont%
      \stuff%
    \endminipage}}%
    \dimen@\ht0
    \advance\dimen@\dp0
  }
  \def\stuff{\BODY}
  \buildbox
  \ifdim\dimen@>#2
    \loop
      \fontdim.5\fontdim 
      \buildbox
    \ifdim\dimen@>#2 \repeat
    \lowerfontdim\fontdim
    \upperfontdim2\fontdim
    \fontdim1.5\fontdim
  \else
    \loop
      \fontdim2\fontdim 
      \buildbox
    \ifdim\dimen@<#2 \repeat
    \upperfontdim\fontdim
    \lowerfontdim.5\fontdim
    \fontdim.75\fontdim
  \fi
  \loop
    \buildbox
    \ifdim\dimen@>#2
      \moreiterationstrue
      \upperfontdim\fontdim
      \advance\fontdim\lowerfontdim
      \fontdim.5\fontdim
    \else
      \advance\dimen@-#2
      \ifdim\dimen@<10pt
        \lowerfontdim\fontdim
        \advance\fontdim\upperfontdim
        \fontdim.5\fontdim
        \dimen@\upperfontdim
        \advance\dimen@-\lowerfontdim
        \ifdim\dimen@<.2pt
          \moreiterationsfalse
        \else
          \moreiterationstrue
        \fi
      \else
        \moreiterationsfalse
      \fi
    \fi
  \ifmoreiterations \repeat
  \box0
}
\makeatother

\title{Investigating Code-Mixed Modern Standard Arabic-Egyptian to English Machine Translation }

\author{ El Moatez Billah Nagoudi ~~~ AbdelRahim Elmadany  ~~~Muhammad Abdul-Mageed  \\
\normalsize Natural Language Processing Lab  \\
  \normalsize The University of British Columbia\\
      
  \texttt{ \small \{moatez.nagoudi,a.elmadany,muhammad.mageed\}@ubc.ca}
  }

\begin{document}
\maketitle
\setcode{utf8}

\input{abstract}

\input{intro}

\input{related_work}
\input{STask}

\input{Para_Data}

\input{MT_Models}

\input{Eval}

\input{official}

\input{discussion}
\input{Concl}

\bibliography{anthology,custom}
\bibliographystyle{acl_natbib}

\end{document}

%% file: abstract.tex
\section*{~~~~~~~~~~~~~~~~~~~~~~~~~~~~~Abstract}

\begin{fitbox}{0.46\textwidth}{0.47\textheight}

Recent progress in neural machine translation (NMT) has made it possible to translate successfully between monolingual language pairs where large parallel data exist, with pre-trained models improving performance even further. Although there exists work on translating in code-mixed settings (where one of the pairs includes text from two or more languages), it is still unclear what recent success in NMT and language modeling exactly means for translating code-mixed text. We investigate one such context, namely MT from code-mixed Modern Standard Arabic and Egyptian Arabic (MSAEA) into English. We develop models under different conditions, employing both (i) standard end-to-end sequence-to-sequence (S2S) Transformers trained from scratch and (ii) pre-trained S2S language models (LMs). We are able to acquire reasonable performance using only MSA-EN parallel data with S2S models trained from scratch. We also find LMs fine-tuned on data from various Arabic dialects to help the MSAEA-EN task. Our work is in the context of the Shared Task on Machine Translation in Code-Switching. Our best model achieves $\bf25.72$ BLEU, placing us first on the official shared task evaluation for MSAEA-EN.  \\
\end{fitbox}

%% file: intro.tex
\section{Introduction}
Recent year have witnessed fast progress in various areas of natural language processing (NLP), including machine translation (MT) where neural approaches have helped boost performance when translating between pairs with especially large amounts of parallel data. However, tasks involving a need to process data from different languages mixed together remain challenging for all NLP tasks. This phenomenon of using two or more languages simultaneously in speech or text is referred to as \textit{code-mixing}~\cite{gumperz_1982} and is prevalent in multilingual societies~\cite{sitaram_survey19}. Code-mixing is challenging since the space of possibilities when processing mixed data is vast, but also because there is not usually sufficient code-mixed resources to train models on. Nor is it clear how much code-mixing existing language models may have seen during pre-training, and so ability of these language models to transfer knowledge to downstream code-mixing tasks remain largely unexplored.  
\input{Tabels/MT_DA_challenges}
In this work, we investigate translation under a code-mixing scenario where sequences at source side are a combination of two varieties of the collection of languages referred to as Arabic. More specifically, we take as our objective translating between Modern Standard Arabic (MSA) mixed with Egyptian Arabic (EA) (source; collectively abbreviated here as MSAEA) into English (target). Table\ref{MSADA-ENG_tran} shows two examples of MSAEA sentences and their human and machine translations. We highlight problematic translations caused by nixing of Egyptian Arabic with MSA.  Through work related to the shared task, we target the following three main research questions:

\begin{enumerate}
    \item \textit{How do models trained from scratch on purely MSA data fare on the code-mixed MSAEA data (i.e., the \textit{zero-shot EA} setting)?} 
    \item \textit{How do existing language models perform under the  code-mixed condition (i.e., MSAEA)?} 
    \item \textit{What impact, if any, does exploiting dialectal Arabic (DA) data (i.e., from a range of dialects) have on the MSAEA code-mixed MT context?} 
\end{enumerate} 

\noindent Our main contributions in this work lie primarily in answering these three questions. We also develop powerful models for translating from MSAEA to English.


The rest of the paper is organized as follows: Section~\ref{sec:relwork} discusses related work. The shared task is described in Section~\ref{sec:shtask}. Section ~\ref{sec:data} describes external parallel data we exploit to build our models. Section~\ref{sec:our_MT_models} presents the  proposed  MT  models. Section~\ref{sec:ourappr} presents our experiments, and our different settings. We provide evaluation on Dev data in Section~\ref{sec:eval_on_DEV} and official results in Section~\ref{sec:official_res}. We conclude in Section~\ref{sec:conclusion}.

%% file: Tabels/MT_DA_challenges.tex
\begin{table}[t]
\centering
 \renewcommand{\arraystretch}{1.5}
\resizebox{1\columnwidth}{!}{%
   
        \begin{tabular}{lll}
        \toprule
&        \textbf{\textbf{(1)} \small{  MSAEA }} & ~~~~~~~~~~~~~~.\small \< أنا عايز شغل جامد يا جدعان> \\\toprule
  \multirow{3}{*}{\rotatebox[origin=c]{90}{\textbf{\colorbox{blue!10}{\small English}}}}     &       \textbf{\small{Human} }   & \colorbox{green!10}{I want hard work, guys.}  \\   \cdashline{2-3}
 &      \textbf{\small{GMT} }     & \colorbox{green!10}{I want a} \colorbox{red!10}{rigid} \colorbox{green!10}{job,} \colorbox{red!10}{Jadaan.}  \\  \cdashline{2-3}
    &         \textbf{\small{S2ST} }     & \colorbox{green!10}{I want a} \colorbox{red!10}{solid} \colorbox{green!10}{job,} \colorbox{red!10}{jadan.}  \\
        
               \toprule    
& \textbf{\textbf{(2)}\small{ MSAEA }} & \small~~.\<الدكاترة قالو انى مش همشي طبيعى  تانى  >\\ \toprule
 \multirow{6}{*}{\rotatebox[origin=c]{90}{\textbf{\colorbox{blue!10}{\small  English }}}}  &   \multirow{2}{*}{\bf  \small   {Human}}\textbf{}                     &    \colorbox{green!10}{The doctors said I can't walk} \\
   &            &   \colorbox{green!10}{normally again.}\\ \cdashline{2-3}
 & \multirow{2}{*}{\bf  \small   {GMT}}\textbf{}    &    \colorbox{green!10}{The doctors said that} \colorbox{red!10}{I was not a} \\
   &            &    \colorbox{red!10}{ normal marginal} \colorbox{green!10}{again.} \\  \cdashline{2-3}
 & \multirow{2}{*}{\bf  \small   {S2ST}}   &    \colorbox{green!10}{Doctors said I wasn't a} \colorbox{red!10}{natural}  \\ 
&               &    \colorbox{red!10}{ marginality} \colorbox{green!10}{again.} \\

             \toprule  
        \end{tabular}}
    \caption{Code-mixed  Modern Standard Arabic-Egyptian Arabic (MSAEA)  sentences with their English human  translation, Google machine translation (GMT)\footnote{We use Google Translate API \url{https://cloud.google.com/translate}.}, and translation by a sequence-to-sequence Transformer model (S2ST) trained from scratch on $55$M MSA-English parallel sentences. \colorbox{green!10}{\textbf{Green}} refers to good translations. \colorbox{red!10}{\textbf{Red}} refers to erroneous translation.}
    \label{MSADA-ENG_tran}
\end{table}

%% file: related_work.tex
\section{Related Work}
\label{sec:relwork}
A thread of research on code-mixed MT focuses on automatically generating synthetic code-mixed data to improve the downstream task. This includes attempts to generate linguistically-motivated sequences~\cite{pratapa-etal-2018-language}. Some work leverages sequence-to-sequence (S2S) models ~\cite{ winata-etal-2019-code} to generate code-mixing exploiting an external neural MT system, while others~\cite{garg-etal-2018-code} use a recurrent neural network along with data generated by a sequence generative adversarial network (SeqGAN) and grammatical information such as from a part of speech tagger to generate code-mixed sequences. These methods have dependencies and can be costly to scale beyond one language pair. \\

\noindent\textbf{Arabic MT.} For Arabic, some work has focused on translating between MSA and Arabic dialects. For instance, \newcite{zbib2012machine} studied  the  impact  of  combined  dialectal  and  MSA  data  on dialect/MSA to English MT  performance. \newcite{sajjad2013translating} uses MSA as a pivot language for translating Arabic dialects into English. \newcite{salloum2014sentence} investigate the effect of sentence-level dialect identification and several linguistic features for MSA/dialect-English translation. \newcite{guellil2017neural} propose an neural machine translation (NMT) system for Arabic dialects using a vanilla recurrent neural networks (RNN) encoder-decoder model for translating Algerian Arabic written in a mixture of Arabizi and Arabic characters into MSA. \newcite{baniata2018neural} present an NMT system to translate Levantine (Jordanian, Syrian, and Palestinian) and Maghrebi (Algerian, Moroccan, Tunisia) to MSA, and MSA to English.
\newcite{farhan2020unsupervised}, propose  unsupervised dialectal NMT, where the source dialect is not represented in training data. This last problem is referred to as zero-shot MT \cite{lample2018phrase}. \\ 
\noindent\textbf{\\DA Arabic MT Resources.}
There are also efforts to develop dialectal Arabic MT resources. For example,  
\citet{meftouh-etal-2015-machine} present the Parallel Arabic Dialect Corpus  (PADIC),\footnote{\url{https://sites.google.com/site/torjmanepnr/6-corpus}} which is a multi-dialect corpus including MSA, Algerian, Tunisian, Palestinian, and Syrian. Recently,~\newcite{sajjad2020arabench} also introduced AraBench, an evaluation suite for dialectal Arabic to English MT. AraBench consists of five publicly available datasets: Arabic-Dialect/English Parallel Text (APT) ~\cite{zbib2012machine}, Multi-dialectal Parallel Corpus of Arabic (MDC)~\cite{bouamor2014multidialectal}, MADAR Corpus~\cite{bouamor2018madar}, Qatari-English speech corpus~\cite{elmahdy2014development}, and the English Bible translated into MSA.\footnote{The United Bible Societies https://www.bible.com}

%% file: STask.tex
\section{Code-Switching Shared Task}
\label{sec:shtask}
The goal of the shared tasks on machine translation in code-switching settings\footnote{\url{https://code-switching.github.io/2021}.} is to encourage building MT systems that translate a source sentence into a target sentence while one of the directions  contains an alternation between two languages (i.e., code-switching). We note that, in the current paper, we employ the wider term \textit{code-mixing}. The shared task involves two subtasks: 
\begin{enumerate}
    \item \textbf{Supervised MT.}  For supervised MT, gold data are provided to participants for training and evaluating models that take English as input and generate Hinglish sequences.
    
    \item \textbf{Unsupervised MT. } In this subtask,  the goal is to develop systems that can generate high quality translations  for multiple language combinations. These combinations include  Spanish-English to English or Spanish, English to Spanish-English, Modern Standard Arabic-Egyptian Arabic (MSAEA) to English and vice versa. For each pair, only test data are provided to participants, with no reference translations.
\end{enumerate}

\noindent In the current work, we focus on the unsupervised MT subtask only. More specifically, we build models exclusively for MSAEA to English. Our approach exploits external data to train a variety of models. We now describe these external datasets.

%% file: Para_Data.tex
\section{Parallel Datasets}\label{sec:data}
\subsection{MSA-English Data }\label{subsec:MSA-eng}
In order to develop Arabic MT models that can translate efficiently across different text domains, we make use of a large collection of parallel  sentences extracted from the Open Parallel Corpus (OPUS)~\cite{OPUS}. OPUS contains more than $2.7$ billion parallel sentences in $90$ languages. To train our models, we extract more than $\sim 61$M sentences MSA-English parallel  sentences from the whole collection. Since OPUS can have noise and duplicate data, we clean this collection and remove duplicates before we use it. We now describe our quality assurance method for cleaning and deduplication of the data.

\noindent\textbf{Data Quality Assurance.}
To keep only high quality parallel sentences,  we follow two steps: 

\begin{enumerate}
    \item We run a cross-lingual semantic similarity model \cite{yang2019multilingual} on each pair of sentences, keeping only sentences with a bilingual similarity score between $0.30$ and $0.99$. This allows us to filter out sentence pairs whose source and target are identical (i.e., similarity score = $1$) and those that are not good translations of one another (i.e., those with a cross-lingual semantic similarity score $< 0.3$).

    \item Observing some English sentences in the source data, we perform an analysis based on sub-string matching between source and target, using the word trigram sliding window method proposed by \newcite{barron2009automatic} and used in \newcite{abdul2021mega} to de-duplicate the data splits. In other words, we compare each sentence in the source side (i.e., MSA) to the target sentence (i.e., English). We then inspect all pairs of sentences that match higher than a given threshold, considering thresholds between $90\%$ and $30\%$. We find that a threshold of $> 75\%$ safely guarantees completely distinct source and target pairs.

\end{enumerate}

\noindent More details about the MSA-English OPUS dataset before and after our quality assurance, including deduplication, are provided in Table~\ref{tab:tab-opus}.
\input{Tabels/opus_data}

\subsection{Dialectal Arabic-English Data}\label{subsec:daiadata}

Several recent works show that MT models trained on one dialect can be used to improve models targeting other dialects~\cite{farhan2020unsupervised, sajjad2020}. For this reason, we exploit several parallel dialectal Arabic (DA)-English datasets in order to enhance the  MSAEA to English translation. 

\noindent \textbf{DA-English Parallel Corpus.} \newcite{zbib2012machine} provide  $38$k Egyptian Arabic (EA)-English and $138$k Levantine-English sentences ($\sim 3.5$ million tokens of Arabic dialects), collected from online user groups and dialectal Arabic weblogs. The authors use crowdsourcing to translate this dataset into English.

\noindent \textbf{MADAR Corpus. }  MADAR \newcite{bouamor2018madar} is a commissioned dataset where $26$ Arabic native speakers were tasked to translate $2$k English sentences each into their own native dialect. In addition,~\newcite{bouamor2018madar} translate $10$k more sentences for five selected cities: Beirut, Cairo, Doha, Cairo, Tunis, and Rabat. The MADAR dataset also has region-level categorization (i.e., Gulf, Levantine, Nile, and Maghrebi). In our work, we use only the Gulf, Levantine,  and Nile (Egyptian) dialects, and exclude Maghrebi.\footnote{We do not make use of the Maghrebi data due to the considerable linguistic differences between Maghrebi and the the Egyptian dialect we target in this work.}  

\noindent \textbf{Qatari-English Speech Corpus.} This parallel corpus comprises $14.7$k Qatari-English sentences collected  by~\newcite{elmahdy2014development} from talk-show programs and Qatari TV series. 

\noindent More details about all our  parallel dialectal-English datasets are in Table~\ref{tab:dia}.
\input{Tabels/dia_data}

%% file: Tabels/opus_data.tex
\begin{table}[t]
\centering
 \renewcommand{\arraystretch}{1.1}
\resizebox{1\columnwidth}{!}{%
\begin{tabular}{lr}\toprule
      \textbf{ Data}                        &\textbf{ \#Sentences}  \\\toprule
{Bible}             & $62.2$K                                       \\
{EUbookshop}             & $1.7$K                                        \\
{GlobalVoices}          & $52.6$K                                       \\
{Gnome}                     & $150$                                          \\
{Infopankki}                & $50.8$K                                       \\
{KDE4}                      & $116.2$K                                      \\
{MultiUN}                   & $9.8$M                                    \\
{News Commentary}           & $90.1$K                                      \\
{OpenSubtitles}             & $29.8$M                                   \\
{QED}                       & $500.9$K                                      \\
{Tanzil}                    & $187$K                                      \\
{Tatoeba}                   & $27.3$K                                       \\
{TED2013}                   & $152.8$K                                      \\
{Ubuntu}                    & $6$K                                        \\
{UN}                        & $74.1$K                                       \\
{UNPC}                      & $20$M                                   \\
{Wikipedia}                 & $151.1$K                                      \\  \toprule   
\textbf{Total}                     & $61$M                         \\ \hline
\textbf{Similarity $\in$  {[}$0.3$ - $0.99${]}}      & $5.7$M             \\ \hline
\textbf{\textit{N-}gram deduplication (\textgreater{}$0.75$)} & $55.2$M  \\      \toprule                 
\end{tabular}}
\caption{Parallel datasets extracted from OPUS \cite{OPUS}. We remove duplicate and identical pairs, keeping only high quality translations. }
\label{tab:tab-opus}

\end{table}

%% file: Tabels/dia_data.tex
\begin{table}[]
\centering
 \renewcommand{\arraystretch}{1.5}
\resizebox{1\columnwidth}{!}{%
\begin{tabular}{lccc}
\toprule
    \textbf{Dataset}   & \textbf{Egyptian}                   & \textbf{Levantine}            & \textbf{Gulf}                  \\
\toprule
\newcite{bouamor2018madar} &  $18$K  & $22$K &  $26$K                   \\
\newcite{elmahdy2014development}    & $-$  & $-$ &  $14.7$K   \\
\newcite{zbib2012machine} & $38$K            & $138$K  & $-$ \\

\toprule
\textbf{Total}   & $56$K  & $160$K&  $40.7$K   \\
\toprule
\end{tabular}}
\caption{Our parallel  DA-English datasets. Gulf comprises data from \newcite{bouamor2018madar}, \newcite{elmahdy2014development} , and \newcite{zbib2012machine}.}
\label{tab:dia}

\end{table}

%% file: MT_Models.tex
\subsection{Data Splits and Pre-Processing}\label{subsec:splits_preproc}
\textbf{Data Splits.} For our experiments, we split the MSA and DA data as follows:

\noindent\textbf{MSA.} We randomly pick $10$k sentences for validation (MSA-Dev) from MSA parallel data (see Section~\ref{subsec:MSA-eng}) after cleaning, and we use the rest of this data ($\sim55.14$M) for training (MSA-Train).
\input{Tabels/Examples.tex}
\noindent\textbf{DA.} For validation (DA-Dev), we randomly pick $6$k sentences  from the $38$k  Egyptian-English data provided by  \newcite{zbib2012machine}. We then use the rest of the data  (i.e., $\sim250.7$k) for training (DA-Train). 


\noindent\textbf{Pre-Processing.} Pre-processing is an important step for building any MT model as it can  significantly affect end results~\cite{oudah2019impact}. For all our models, we only perform light pre-processing in order to retain a faithful representation of the original (naturally occurring) text. We remove diacritics and replace URLs, user mentions, and hashtags with the generic string tokens \texttt{URL}, \texttt{USER}, and \texttt{HASHTAG} respectively. Our second step for pre-processing is specific to each type of models we train as we will explain in the respective sections.

\section{MT Models}\label{sec:our_MT_models}
\subsection{From-Scratch Seq2Seq Models}\label{subsec:Seq2Seq}
We train our models on the MSA-English parallel data  described in section~\ref{subsec:MSA-eng} on MSA-Train with a Transformer~\cite{vaswani2017attention} model as implemented in Fairseq~\cite{ott2019fairseq}. For that, we follow \citet{ott2018scaling} in using $6$ blocks for each of the encoder and decoder parts. We use a learning rate of $0.25$, a dropout of $0.3$, and a batch size  $4,000$ tokens. For the optimizer, we use Adam \cite{kingma2014adam} with beta coefficients of $0.9$ and $0.99$ which control an exponential decay rate of running averages, with a weight decay of $10^{-4}$.  We also apply an inverse square-root learning rate scheduler with a value of $5e^{-4}$ and $4,000$ warm-up updates. For the loss function, we use label smoothed cross entropy with a smoothing strength of $0.1$. We run the 
Moses tokenizer \cite{koehn2007open} on our input before passing data to the model. For vocabulary, we use a joint Byte-Pair Encoding (BPE)~\cite{sennrich2015neural} vocabulary with $64$K split operations for subword segmentation.


\subsection{Pre-Trained Seq2Seq Language Models} 
\label{subsec:Pre-Trained}
We also fine-tune two state-of-the-art pre-trained multlingual generative models, mT5~\cite{xue2020mt5} and mBART~\cite{liu2020multilingual} on DA-Train  for $100$ epochs. We use early stopping during fine-tuning and identify the best model on DA-Dev. We use the HuggingFace~\cite{wolf2020transformers} implementation of each of these models, with the default settings for all hyper-parameters.

\section{Experiments and  Settings}\label{sec:ourappr}

In this section, we describe the different ways we fine-tune and evaluate  our models.

\subsection{Zero-Shot Setting}  First, we use S2ST model trained on MSA-English data exclusively to evaluate  MSAEA code-mixed data . While we can refer to this setting as \textit{zero-shot}, we note that it is not truly zero-shot in the strict sense of the word due to the code-mixed nature of the data (i.e., the data has a mixture of MSA and EA). Hence, we will refer to this setting as \textit{zero-shot EA}.

\subsection{Fine-Tuning Setting}
Second, we further fine-tune the three models (i.e., S2ST, mT5, and mBART) on the DA data described in Section~\ref{subsec:daiadata}. While the downstream shared task data only involves EA mixed with MSA, we follow~\newcite{farhan2020unsupervised} and ~\newcite{sajjad2020} in fine-tuning on different dialects when targeting a single downstream dialect (EA in our case). We will simply refer to this second setting as \textit{Fine-Tuned DA}.

%% file: Tabels/Examples.tex
\begin{table*}[ht]
\centering
 \renewcommand{\arraystretch}{1.6}
\resizebox{\textwidth}{!}{%
\begin{tabular}{ll}
\toprule 

\textbf{Source:} & ~~~~~~~~~~~~~~~~~~~~~~~~~~~~~~~~~~~~~~~~~~~~~~~~~~~~~~~~~~~~~~~~~~~~~~~~~~~~~~~~~~~~~~~~~~~~~~~~~~~~~~~~~~~~~~~.\< مش عارفين نتأكد و مش عارفين البنات فين   >  \\
\cline{2-2}
\textbf{S2ST} & \colorbox{green!10}{we don't know for sure and the girls don't know} \colorbox{red!10}{finn}.                                      \\
\cdashline{2-2}

\textbf{mT5}    & \colorbox{green!10}{we can't make sure and we don't know where the girls are }                                                                                                                                                             \\
\cdashline{2-2}
\textbf{mBART}  & \colorbox{red!10}{we don't know where to make sure} \colorbox{green!10}{ and we don't know where the girls are }                                                                                                                                                 \\
\hline
\textbf{Source:} & \begin{tabular}[c]{@{}l@{}}~~~~~\<انا عايز اعرف موقف الاخوان الرسمي من التحرش بالحريري و نجاد البرعي ولو البلطجية دول مش تبعهم الرئيس >\\ 
~~~~~~~~~~~~~~~~~~~~~~~~~~~~~~~~~~~~~~~~~~~~~~~~~~~~~~~~~~~~~~~~~~~~~~~~~~~~~~~~~~~~~~~~~~~~~~~~~~~~~~~~~~~~~~~~~~~~~~~~~~~~~~~~~~~~~~~~~~~~~~~~~~~~~~~~~~~.\< يستعمل سلطته>\end{tabular}                                                       \\\cline{2-2}
\textbf{S2ST} & \begin{tabular}[c]{@{}l@{}}\colorbox{green!10}{i want to know} \colorbox{red!10}{the brothers'}\colorbox{green!10}{ official position on harassment of} \colorbox{red!10}{liberals} \colorbox{green!10}{and nejad al-barai, even the } \colorbox{red!10}{thugs, } \\\colorbox{red!10}{countries} \colorbox{green!10}{that are not followed by the president are using his authority and ordering their immediate arrest.} \end{tabular}\\\cdashline{2-2}

\textbf{mT5}    & \begin{tabular}[c]{@{}l@{}}\colorbox{green!10}{i want to know the situation of   the official} \colorbox{red!10}{brothers} \colorbox{green!10}{from harassment of} \colorbox{red!10}{the silky} \colorbox{green!10}{and najad }\colorbox{red!10}{albarea} \colorbox{green!10}{and }\\\colorbox{red!10}{if   these pants }\colorbox{green!10}{are not their president the president uses his power and order to   arrest them immediately}\end{tabular}  \\
 \cdashline{2-2}
\textbf{mBART}  & 	\begin{tabular}[c]{@{}l@{}}\colorbox{green!10}{i want to know the position of   the official} \colorbox{red!10}{brothers} \colorbox{green!10}{from harassment in} \colorbox{red!10}{the army} \colorbox{green!10}{and najad al-bara'y,} \\\colorbox{red!10}{even   if these are not theirs}\colorbox{green!10}{, the president should use his authority and order to   arrest them immediately }  \end{tabular} \\

\hline

\textbf{Source:} & ~~~~~~~~~~~~~~~~~~~~~~~~~~~~~~~~~~~~~~~~~~~~~~~~~“ user : \<عاوزين محامي يروح معانا القسم يا جدعان صبحي ووليد تليفوناتهم مقفولة >.”                                                
\\\cline{2-2}

\textbf{S2ST} & user: \colorbox{green!10}{there is a need for a lawyer} \colorbox{red!10}{to help the section, jadan} \colorbox{green!10}{sobhi and walid} \colorbox{red!10}{televonas} \colorbox{green!10}{closed}. \\\cdashline{2-2}

\textbf{mT5}    & user: \colorbox{green!10}{we want a lawyer to go   with us to }\colorbox{red!10}{the section}, \colorbox{green!10}{guys}, \colorbox{green!10}{sobhe and waleed their telephones are   closed}                                                                                                              \\\cdashline{2-2}
\textbf{mBART}  & << user : \colorbox{green!10}{we want a lawyer}   \colorbox{green!10}{to go with us to} \colorbox{red!10}{the section,} \colorbox{red!10}{oh good morning, and}  \colorbox{green!10}{their telephones are   closed.} >>                                                                                                         \\

\hline 
\textbf{Source} &  ~~~~~~\begin{tabular}[c]{@{}l@{}}\<بيعقدوا الجلسات في أماكن مش محاكم , و ما ينفعش خلق الله يدخلوا من غير تصريح , عشان المتهمين لما ييجوا >\\ 
~~~~~~~~~~~~~~~~~~~~~~~~~~~~~~~~~~~~~~~~~~~~~~~~~~~~~~~~~~~~~~~~~~~~~~~~~~~~~~~~~~~~~~~~~~~~~~~~~~~~~~~~~~~~~~~~~~~~~~~~~~~~~~~~~~~~~\<يمنعوهم و يحكموا غيابي !!   >\end{tabular}  \\
\cline{2-2}

\textbf{S2ST}  & \begin{tabular}[c]{@{}l@{}}\colorbox{red!10}{they hold hearings} \colorbox{green!10}{in places where there are no courts, and} \colorbox{red!10}{what thrives on god's creation} \colorbox{green!10}{will enter without}\\ \colorbox{green!10}{permission, because the accused} \colorbox{red!10}{will not prevent them and judge my absence!}\end{tabular}\\ \cdashline{2-2}
\textbf{mT5}    & \begin{tabular}[c]{@{}l@{}}\colorbox{green!10}{they have sessions in places   that are not courts, and} \colorbox{red!10}{god doesn't allow people }\colorbox{green!10}{to enter without a permit, so that } \\\colorbox{green!10}{when they come and prevent them and }\colorbox{red!10}{rule me absence }                                              \end{tabular}  \\\cdashline{2-2}
\textbf{mBART}  & \begin{tabular}[c]{@{}l@{}}\colorbox{green!10}{they hold meetings in places where there is no courts, and} \colorbox{red!10}{god doesn't allow people} \colorbox{green!10}{to enter without a permit, }\\ \colorbox{green!10}{so that when the} \colorbox{red!10}{accused come they stop them and rule them }                                           \end{tabular}  \\

\toprule 
\end{tabular}%
}
\caption{MSA-EA sentences with their English translations using our Models. \textbf{S2ST:} Sequence-to-sequence Transformer model trained from scratch. Data samples are extracted from the shared task Test data. \colorbox{green!10}{\textbf{Green}} refers to good translation. \colorbox{red!10}{\textbf{Red}} refers to problematic translation.}
    \label{tab:results_examples}
\end{table*}

%% file: Eval.tex
\section{Evaluation on Dev Data}\label{sec:eval_on_DEV}

We report results of all our models under different settings in BLEU scores~\cite{papineni2002bleu}.
In addition to evaluation on uncased data, we run a language modeling based truecaser~\cite{lita2003truecasing} on the outputs of our different models.\footnote{We were not been able to report results based on truecasing in this paper, but we note that we will provide these results in the camera ready version of this paper.} Results presented in Table~\ref{tab:devres1} show that S2ST achieves relatively low scores (between $8.54$ and $12.57$) on all settings. In comparison, both mBART and mT5 fine-tuned on DA-Train are able to translate MSAEA to English with BLEU scores of $23.80$ and $24.70$ respectively. We note that truecasing the output results in improving the  results with an average of $+2.55$ BLEU points.
\input{Tabels/dev_results}

%% file: Tabels/dev_results.tex
\begin{table}[t]
\centering
 \renewcommand{\arraystretch}{1.3}
\resizebox{1\columnwidth}{!}{%
\small 
\begin{tabular}{clc}
\toprule
    \textbf{Model}   & \textbf{Setting}                   & \textbf{BLEU}          \\
\toprule
\multirow{4}{*}{\bf  \small \colorbox{blue!10}{S2ST}}\textbf{}  &  Zero Shot   EA       &   $8.54$  \\

&  Fine-tuned  DA    &  $9.33$       \\
 &    Zero Shot EA (true-cased)           &  $11.59$  \\ 
 &   Fine-tuned DA (true-cased)           &    $12.57$ \\ \hline
 

\multirow{2}{*}{\bf  \small  \colorbox{blue!10}{mT5}}\textbf{}  &  Fine-tuned  DA      &    $24.70$     \\
 &   Fine-tuned  DA (true-cased)          & $26.35$ \\ \hline


\multirow{2}{*}{\bf  \small  \colorbox{blue!10}{mBART}}\textbf{}  &  Fine-tuned   DA     &         $23.80$ \\
 &   Fine-tuned  DA  (true-cased)           & \colorbox{green!10}{$\bf26.07$}\\ \hline

\toprule
\end{tabular}}
\caption{Results of models on DA-Dev data. \textbf{S2ST:} Sequence-to-sequence Transformer model trained from scratch. We note that in the zero-shot EA setting the S2ST model is trained on $55$M bitext sentences.}
\label{tab:devres1}

\end{table}

%% file: official.tex
\section{Official Shared Task (Test) Results}\label{sec:official_res}
Table~\ref{tab:res1} shows results of all our MT models with different settings on the official shared task Test set. We observe that the Transformer model in the \textit{zero shot EA} setting (a model that does not see Egyptian Arabic data) was able to translate MSAEA to English with $21.34$ BLEU. As expected, fine-tuning all the models on DA-Train improves results across all models and leads to the best BLEU score of $25.72\%$ with the S2ST model. 

Comparing performance of the S2ST model on Dev and Test data, we observe that Test data results are better. This suggests that Test data comprises more MSA than EA sequences. To test this hypothesis, we run a binary MSA-DA classifier~\newcite{abdul2020arbert} on both the Dev and Test data to acquire MSA and DA distributions on each dataset. Results of this analysis, shown in Table~\ref{tab:distr}, confirm our hypothesis about Test data involving significantly more MSA (i.e., $72.31\%$) compared to Dev data.


\input{Tabels/data_dist}

\input{Tabels/results}


%% file: Tabels/data_dist.tex
\begin{table}[t]
\centering
\renewcommand{\arraystretch}{1.2}
\resizebox{0.9\columnwidth}{!}{%
\begin{tabular}{lccc}
\toprule
    \textbf{Dataset}   & \textbf{\#Size}                   & \textbf{MSA}            & \textbf{DA}                  \\
\toprule
\textbf{DA-Dev} &      $6,164$      & $18.36$\% &  $81.64$\%  \\
\textbf{Official Test} &  $6,500$  & $72.31$\% &  $27.69$\%                     \\

\toprule
\end{tabular}}
\caption{The data distribution (MSA Vs DA) in the DA-Dev and the official Test set.}
\label{tab:distr}

\end{table}

%% file: Tabels/results.tex
\begin{table}[h]
\centering
 \renewcommand{\arraystretch}{1.3}
\resizebox{1\columnwidth}{!}{%
\small 
\begin{tabular}{clc}
\toprule
    \textbf{Model}   & \textbf{Setting}                   & \textbf{BLEU}          \\
\toprule
\multirow{4}{*}{\bf  \small  \colorbox{blue!10}{S2ST}}\textbf{}  &  Zero Shot   EA            & $21.34$ \\

&  Fine-tuned  DA    &        $22.51$ \\
 &    Zero Shot EA (true-cased)           & $23.68$ \\ 
 &   Fine-tuned DA (true-cased)           & \colorbox{green!10}{$\bf25.72$} \\ \hline
 

 \multirow{2}{*}{\bf  \small  \colorbox{blue!10}{mT5}}\textbf{}  &  Fine-tuned  DA      &        $16.41$ \\
 &   Fine-tuned  DA (true-cased)          & $18.80$ \\ \hline


\multirow{2}{*}{\bf  \small  \colorbox{blue!10}{mBART}}\textbf{}  &  Fine-tuned   DA     &        $17.17$ \\
 &   Fine-tuned  DA  (true-cased)           & $19.79$ \\ \hline

\toprule
\end{tabular}}
\caption{Results of our models on official Test data. Again, in the zero-shot EA setting the S2ST model is trained on $55$M bitext sentences,}
\label{tab:res1}

\end{table}


%% file: discussion.tex
\noindent\textbf{\\Discussion.} We inspect output translations from our models on Test data. We observe that even though S2ST performs better than the two language models on Test data, both of these models are especially able to translate Egyptian Arabic tokens such as \<فين> in example (1) in Table~\ref{tab:results_examples} well. Again, Test data contain more MSA than DA as we explained earlier and hence the S2ST model (which is trained on $55$M sentence pairs) outperforms each of the two language models. This analysis suggests that fine-tuning the language models on more MSA-ENG should result in better performance.

Returning to our three main research questions, we can reach a number of conclusions. For \textbf{RQ1}, we observe that models trained from scratch on purely MSA data fare reasonably well on the code-mixed MSAEA data (i.e., \textit{zero-shot EA} setting). This is due to lexical overlap between MSA and EA. For \textbf{RQ2}, we also note that language models such as mT5 and mBART do well under the code-mixed condition, more so than models trained from scratch when inference data involve more EA. This is the case even though these language models in our experiments are fine-tuned with significantly less data (i.e., $\sim 250$K pairs) than the from-scratch S2ST models (which are trained on $55$M MSA + $250$K DA pairs). For \textbf{RQ3}, our results show that training on data from various Arabic dialects helps translation in the MSAEA code-mixed condition. This is in line with previous research~\cite{farhan2020unsupervised} showing that exploiting data from various dialects can help downstream translation on a single dialect dialect in the zero-shot setting.




%% file: Concl.tex
\section{Conclusion}\label{sec:conclusion}
We described our contribution to the  shared tasks on MT in code-switching.\footnote{https://code-switching.github.io/2021} Our models target the MSAEA to English task under the unsupervised condition. Our experiments show that training models on MSA data is useful for the MSAEA-to-English task in the zero-shot EA setting. We also show the utility of pre-trained language models such as mT5 and mBART on the code-mixing task. Our models place first in the official shared task evaluation. In the future, we intend to apply our methods on other dialects of Arabic and investigate other methods such as backtranslation for improving overall performance.


\vspace{2mm}
\section*{Acknowledgements}\label{sec:acknow}

\vspace{2mm}
We gratefully acknowledges support from the Natural Sciences and Engineering Research Council of Canada, the Social Sciences and Humanities Research Council of Canada, Canadian Foundation for Innovation, Compute Canada (\url{www.computecanada.ca}) and UBC ARC-Sockeye (\url{https://doi.org/10.14288/SOCKEYE}) and Penguin Computing POD™ (\url{pod.penguincomputing.com}).